\documentclass[10pt,twocolumn,letterpaper]{article}

\usepackage[pagenumbers]{iccv} 

\usepackage{xcolor}
\usepackage{soul}
\definecolor{yunze}{RGB}{255, 0, 0}  

\usepackage{color, colortbl}

\newcommand{\cmark}{\ding{51}}%
\newcommand{\xmark}{\ding{55}}%

\definecolor{Gray}{gray}{0.5}
\definecolor{LGray}{gray}{0.9}
\definecolor{darkblue}{RGB}{94,110,186}
\definecolor{darkGreen}{RGB}{92, 148, 110}
\definecolor{myblue}{RGB}{14, 121, 178}
\definecolor{myred}{RGB}{192, 0, 0}

\usepackage{pifont}

\usepackage[T1]{fontenc}
\definecolor{iccvblue}{rgb}{0.21,0.49,0.74}
\usepackage[pagebackref,breaklinks,colorlinks,allcolors=iccvblue]{hyperref}
\usepackage{multirow}
\def\paperID{13769} 
\def\confName{ICCV}
\def\confYear{2025}

\title{VideoMAP: Toward Scalable Mamba-based Video Autoregressive Pretraining}

\author{    
    Yunze Liu$^{1,2}$\thanks{Equal contribution.} \ \  
    Peiran Wu$^{4}$\footnotemark[1] \ \thanks{Project Leader.} \ 
    Cheng Liang$^{5}$ \ 
    Junxiao Shen$^{4}$\ 
    Limin Wang$^{5,3}$ \ 
    Li Yi$^{1,3,2}$\thanks{Corresponding Author.}
    \\
    \textsuperscript{1}IIIS, Tsinghua University \quad
    \textsuperscript{2}Shanghai Qi Zhi Institute  \quad
    \textsuperscript{3}Shanghai Artificial Intelligence Laboratory \\
    \textsuperscript{4}University of Bristol  \quad  
    \textsuperscript{5}Nanjing University
}

\begin{document}

\maketitle

\begin{abstract}
Recent Mamba-based architectures for video understanding demonstrate promising computational efficiency and competitive performance, yet struggle with overfitting issues that hinder their scalability.
To overcome this challenge, we introduce VideoMAP, a Hybrid Mamba-Transformer framework featuring a novel pre-training approach. VideoMAP uses a 4:1 Mamba-to-Transformer ratio, effectively balancing computational cost and model capacity. This architecture, combined with our proposed frame-wise masked autoregressive pre-training strategy, delivers significant performance gains when scaling to larger models. Additionally, VideoMAP exhibits impressive sample efficiency, significantly outperforming existing methods with less training data. 
Experiments show that VideoMAP outperforms existing models across various datasets, including Kinetics-400, Something-Something V2, Breakfast, and COIN. 
Furthermore, we demonstrate the potential of VideoMAP as a visual encoder for multimodal large language models, highlighting its ability to reduce memory usage and enable the processing of longer video sequences. The code is open-source at \href{https://github.com/yunzeliu/MAP}{https://github.com/yunzeliu/MAP}.
\end{abstract}

\section{Introduction}
\label{sec:intro}

\begin{figure}
    \centering
    \includegraphics[width=\linewidth]{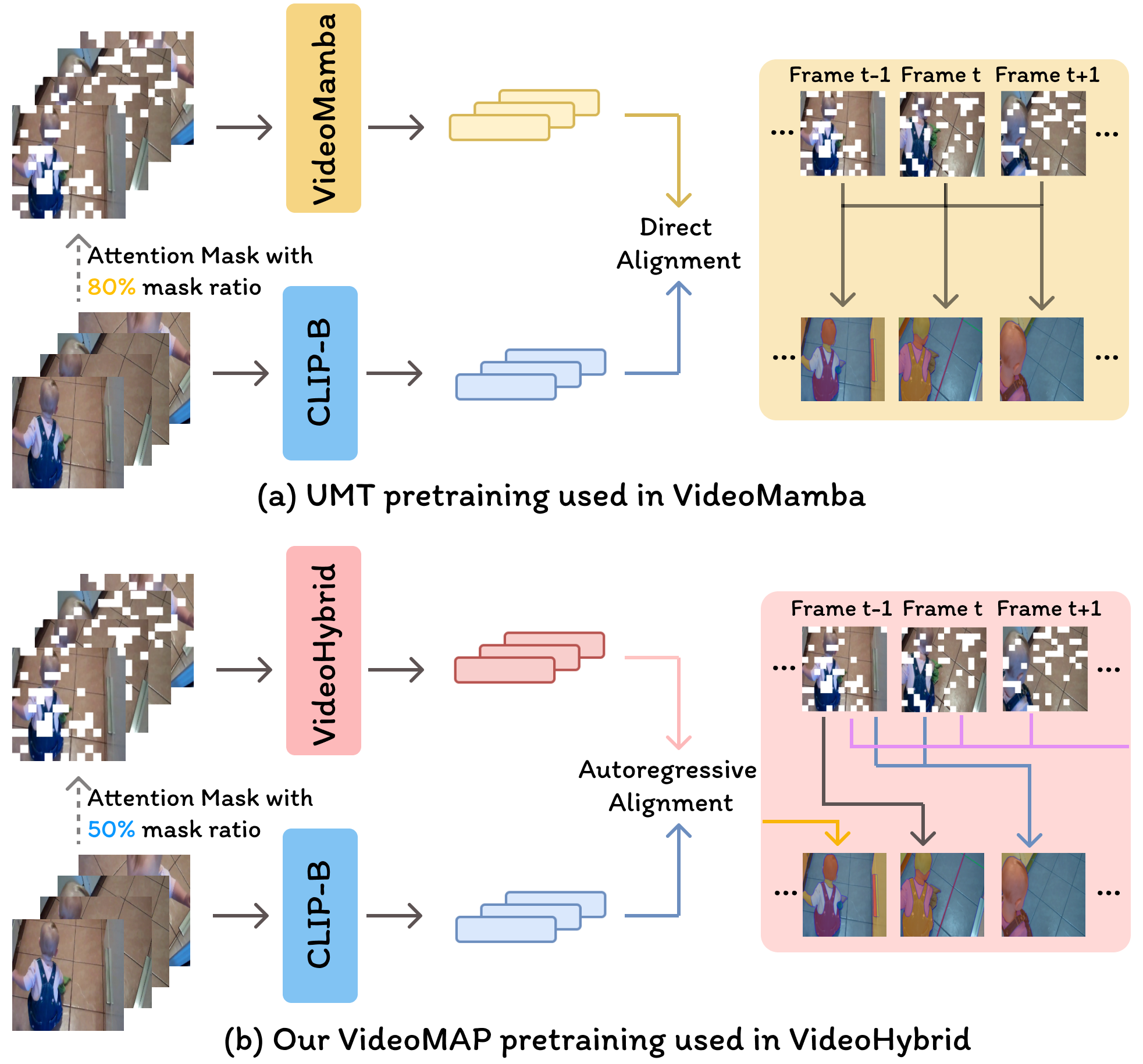}
    \caption{We propose VideoMAP for scalable Mamba-based video autoregressive pretraining. We propose to use a hybrid Mamba-Transformer network and a frame-wise autoregressive prediction method demonstrating stronger scalability and sample efficiency. Unlike the UMT\cite{li2023unmasked} used in VideoMamba\cite{li2025videomamba}, our proposed VideoMAP requires the network to predict both the spatial and semantic information of future frames simultaneously.}
    \vspace{-5mm}
    \label{fig:teaser}
\end{figure}

Video understanding is a crucial task that has consistently attracted significant attention within the research community. Due to the inherent redundancy in video data, researchers strive to develop efficient and effective frameworks for understanding video. A pivotal challenge in video understanding remains balancing computational efficiency with effective processing of extended frame sequences to enhance spatiotemporal representations and downstream performance.

Recently, the emergence of VideoMamba \cite{li2025videomamba} marked a significant milestone in this field, demonstrating impressive performance by effectively modeling longer temporal sequences through the Mamba architecture. However, its scalability and sample efficiency remain unsatisfactory. Specifically, VideoMamba struggles with overfitting and cannot effectively scale beyond 100M parameters. Furthermore, the direct application of pre-training methods designed for Transformers to VideoMamba results in low sample efficiency, requiring substantial data and extended training times to achieve satisfactory performance.

To overcome these challenges, our key finding is that employing a Hybrid Mamba-Transformer framework can significantly mitigate overfitting while achieving significant performance improvements. Additionally, we propose a novel pre-training method specifically designed for the Hybrid Video Mamba-Transformer backbone, which outperforms previous approaches while improving the sample efficiency during pretraining.

Specifically, we introduce VideoMAP, a new hybrid framework for video understanding backbone networks, along with a compatible pre-training strategy. We utilize a 4:1 Mamba-to-Transformer ratio to minimize additional computational overhead while addressing the inherent capacity limitations of Mamba-based architectures, referring to this architecture as VideoHybrid. Building on this, we employ a frame-wise autoregressive pretraining for efficient pretraining of the hybrid framework. Our findings reveal that, unlike the VideoMAE \cite{tong2022videomae} pre-training approach for Transformers, the mask ratio is not a key factor, while frame-wise autoregressive decoding is the most important.

Extensive experiments demonstrate that our approach significantly outperforms existing methods in terms of performance, scalability, and sample efficiency, achieving scalable Mamba-based video understanding. We validated the effectiveness of our method on the K400 \cite{kay2017kinetics}, Something-Something V2 \cite{goyal2017something} datasets, Breakfast\cite{breakfast}, and COIN\cite{coin}. On smaller datasets, our approach yielded even more pronounced improvements, underscoring its sample efficiency. Finally, we also demonstrated the feasibility and strong potential of our framework when applied to VideoLLMs. Compared with pure Mamba and pure Transformer visual encoders, our hybrid visual encoder can significantly reduce GPU memory consumption while maintaining competitive performance.

In summary, the main contribution is threefold: 
\textbf{First}, we propose VideoMAP, a hybrid Mamba-Transformer and frame-wise autoregressive pertaining to achieving scalable Mamba-based video understanding.
\textbf{Second}, extensive experiments and analyses demonstrate the advantages of our VideoMAP in terms of scalability and sample efficiency. On the K400 dataset, the accuracy of VideoMamba\cite{li2025videomamba} was improved from 83.9 to 88.3, significantly improving the capability of Mamba-based models.
\textbf{Third}, we highlight the application potential of the VideoMAP model in VideoLLMs, emphasizing its ability to serve as a visual encoder, which significantly reduces GPU memory usage.

\section{Related Work}
\label{sec:related}
\subsection{Video Understanding}
Video understanding, a pivotal field within computer vision, has seen significant advancements with the integration of multimodal large language models~\cite{chen2024far, lin2023video}. Initially reliant on CNNs for feature extraction from video frames, the field evolved with 3D CNNs~\cite{carreira2017quo, feichtenhofer2020x3d, tran2015learning, tran2018closer} that capture spatial-temporal details. The advent of Transformers, such as TimeSformer~\cite{bertasius2021space} and ViViT~\cite{arnab2021vivit}, introduced attention-based mechanisms that excel in recognizing long-range dependencies in videos. Despite these improvements, computational efficiency remains a challenge. Recent innovations like VideoMamba~\cite{gu2023mamba} in the Mamba-SSM series demonstrate potential in reducing computational demands while enhancing performance. However, it is mentioned that VideoMamba has severe overfitting issues when increasing the number of parameters, which limits its scalability. Our proposed VideoMAP can improve both the model's scalability and performance while maintaining Mamba's efficiency.

\subsection{Self-supervised Video Representation Learning}
Pre-training to enhance generalizability and mitigate overfitting is an important and promising research topic. VideoMAE~\cite{tong2022videomae} adopts Masked Autoencoder (MAE) techniques for videos, highlighting the importance of higher tube-masking strategies and mask ratios. Further developments include VideoMAEv2~\cite{wang2023videomae}, which employs dual masking in the decoder for improved computational and memory efficiencies, and VideoMAC~\cite{pei2024videomac}, which introduces a ConvNet-based video-masked autoencoder. Meanwhile, UMT~\cite{li2023unmasked} focuses on masking low-semantic video tokens while aligning the rest with Image Foundation Models. 
V-JEPA~\cite{bardes2024revisiting} shows that sample efficiency can be achieved by prediction in the latent space instead of pixel reconstruction. Our VideoMAP also demonstrates high sample efficiency.
VideoMamba~\cite{li2025videomamba} integrates these ideas in its pre-training to prevent overfitting. MAP~\cite{liu2024map} framework suggests that traditional Transformer pre-training strategies may not be optimal for Mamba-based models, motivating us to develop VideoMAP. Our VideoMAP uses frame-wise autoregressive alignment for pre-training, enhancing feature alignment compared to VideoMAE by focusing on sequential frame decoding instead of simultaneous multi-frame decoding. Furthermore, unlike MAP and other video pre-training methods, our findings indicate that the mask ratio is not the key factor. Instead, frame-wise autoregressive decoding plays the most critical role during pretraining.

\subsection{Mamba-based Vision Models}
The introduction of the Mamba-SSM model~\cite{gu2023mamba}, with its demonstrated reduction in GPU resource consumption, has paved the way for new opportunities for next-generation unified backbones. In recent years, many researchers have explored whether Mamba-SSM could replace Transformers, leading to the development of models such as Vision Mamba~\cite{zhu2024vision}, VMamba \cite{liu2024vmamba}, Mamba-R~\cite{wang2024mamba}, and VideoMamba~\cite{li2025videomamba}, all of which have demonstrated significant GPU memory savings alongside strong performance.
However, as research into Mamba continues, it has become evident that Mamba's selective mechanism tends to lose certain global details when layers accumulate, and scaling up a purely Mamba-based structure does not lead to significant performance gains,
primarily due to its inherent retrieval capability deficiencies that fundamentally limit its modeling capacity\cite{ShakingupVLMs, CanMambaLearnHowToLearn, RNNs_are_not_Transformers_Yet}. Some studies~\citep{CanMambaLearnHowToLearn,Repeat_After_Me} demonstrate that hybrid architectures integrating SSMs with Transformer components effectively address these limitations.
In response, hybrid Mamba-Transformer architectures have been proposed in the vision domain, showing promising progress~\cite{hatamizadeh2024mambavision, hatamizadeh2024mambavisionhybridmambatransformervision}.
Nevertheless, video understanding typically requires considerable GPU resources. Therefore, a method that not only reduces GPU memory usage but also retains global information during scale-up is important. Our approach aims to address these issues by leveraging a hybrid Mamba-Transformer structure and a novel pretraining paradigm for video applications.

\section{Method}
\label{sec:method}
\begin{figure*}[ht]
    \centering
    \includegraphics[width=1.0\textwidth]{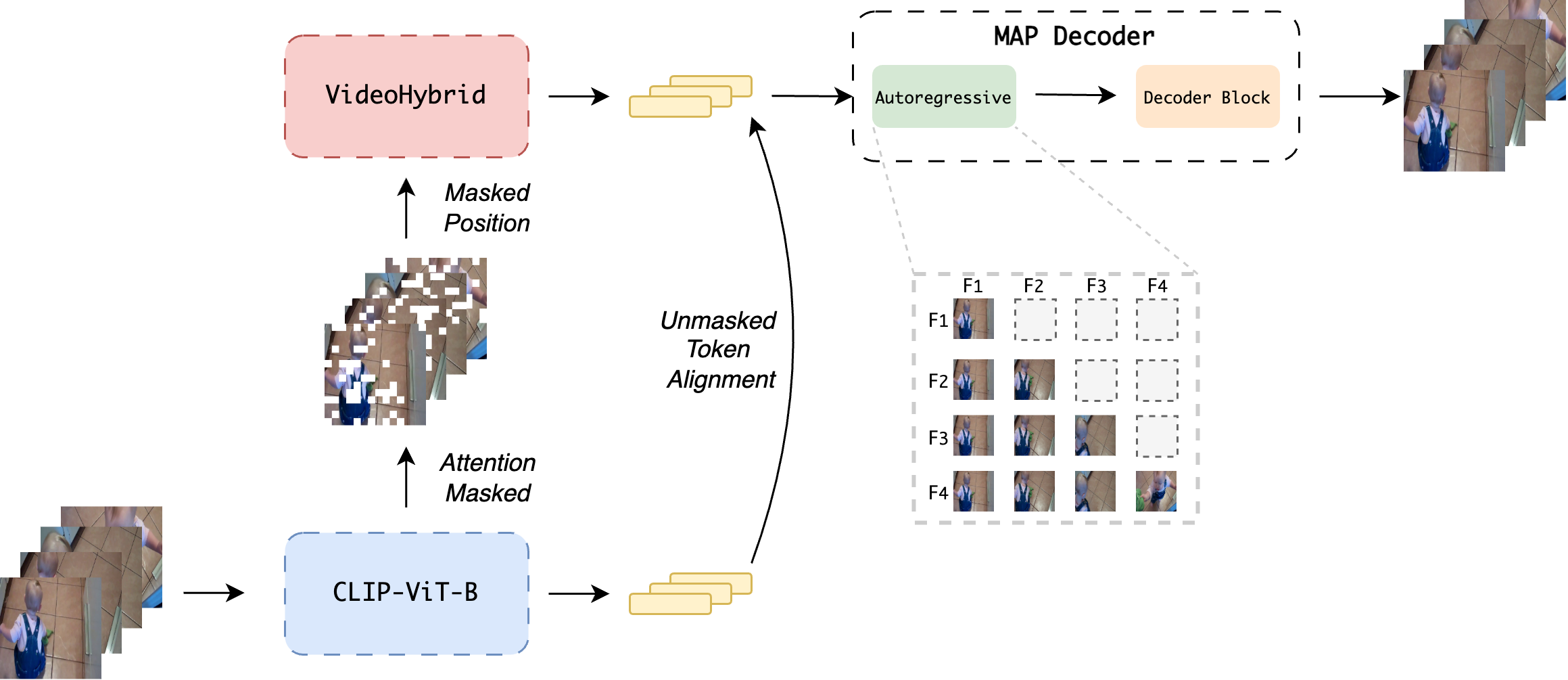} 
    \caption{We propose Video Masked Autoregressive Pretraining (\textbf{VideoMAP}) to pretrain the hybrid Mamba-Transformer vision backbones. This strategy combines the strengths of both UMT and frame-level autoregressive decoder, improving the performance of Transformer and Mamba modules within a unified paradigm.}
    \label{fig:framework_image}
        \vspace{-5mm}
\end{figure*}

We propose a novel video understanding backbone, as illustrated in Figure~\ref{fig:framework_image}. Our approach employs a Hybrid Mamba-Transformer structure as the encoder and introduces a new frame-wise autoregressive decoding strategy. First, we revisit the MAP framework for hybrid Mamba-Transformer for image backbones in Section~\ref{sec:revisit}. We presented an overview of VideoMAP and explained its differences from ImageMAP in Section~\ref{sec:overview}. Then, the encoder strategy is detailed in Section~\ref{sec:Hybrid Mamba-Transformer Video Encoder}, followed by a discussion of the autoregressive decoding strategy in Section~\ref{sec:Autoregressive Decoder}.

\subsection{Revisiting ImageMAP} 
\label{sec:revisit}
The 2D Image Masked Autoencoder Pretraining (ImageMAP) \cite{liu2024map} is an advanced pre-training strategy specifically designed for the Hybrid Mamba-Transformer framework. It has demonstrated superior performance compared to traditional Masked Autoencoder (MAE) and Autoregressive (AR) pre-training methods. The core idea behind ImageMAP is to leverage a combination of masked autoencoding and autoregressive prediction to better capture both spatial and sequential dependencies in images, resulting in more effective feature representations. Its core conclusion is that, compared to MAE\cite{wang2023videomae,tong2022videomae}, a lower mask ratio should be used along with a pre-training order aligned with the Mamba scanning order to achieve optimal performance.

  
  
  

\subsection{Overview of VideoMAP}
\label{sec:overview}
Extending the concept of 2D ImageMAP to video processing, VideoMAP incorporates the spatiotemporal characteristics of videos. Unlike static images, videos contain not only spatial information but also temporal information that captures the relationships between frames. Therefore, we employ frame-wise autoregressive alignment with the Image Foundation Model to more effectively model temporal motion information.

As illustrated in Figure~\ref{fig:framework_image}, VideoMAP is a pre-training strategy for video processing that applies attention masks and prediction across both temporal and spatial dimensions. In contrast to traditional image pre-training methods, VideoMAP uses frame-wise autoregressive pre-training, leveraging the original image features to predict the CLIP features of the subsequent frame.

Given a video \( \mathbf{V} \), we can divide it into \( T \) frames:

\begin{equation}
       \mathbf{V} = \{\mathbf{F}_1, \mathbf{F}_2, \ldots, \mathbf{F}_T\}
\end{equation}

Since Table~\ref{table3} shows that varying the mask ratio has little impact on the final results, we opted not to pursue an extensive design and instead adopted the 50\% ratio used in MAP. Consistent with UMT, we employ an attention mask as our masking strategy. For each frame \( \mathbf{F}_t \), let \(\mathbf{M}_{ti} \subset \{1, 2, \ldots, N\}\) denote the indices of the masked tokens for frame \( \mathbf{F}_t \).

For a masked frame \( \mathbf{F}_t \), the model aims to predict the CLIP features of the subsequent frame. The prediction process utilizes both the visible pixels of the current frame and information from the preceding frames. Specifically, we use the original image features of the current frame \( \mathbf{F}_t \) to predict the CLIP features of the next frame \( \mathbf{F}_{t+1} \):

\begin{equation}
  p(\mathbf{z}_{t+1} \mid \mathbf{F}_t, \mathbf{F}_{<t})
\end{equation}

\noindent where \(\mathbf{z}_{t+1}\) represents the CLIP features of the next frame, and \(\mathbf{F}_{<t}\) denotes all frames preceding frame \( t \).

The overall loss function is defined as the reconstruction error between the CLIP features of the masked frames and their predicted values:

\begin{equation}
  \mathcal{L} = \sum_{t=1}^{T-1} \mathcal{L}_{\text{recon}}(\mathbf{z}_{t+1}, \hat{\mathbf{z}}_{t+1})
\end{equation}

\noindent where \(\mathcal{L}_{\text{recon}}\) is the reconstruction loss function, and \(\hat{\mathbf{z}}_{t+1}\) represents the predicted CLIP features of the next frame.

The key differences between our VideoMAP and ImageMAP are reflected in three aspects. 
\begin{itemize}
    \item First, we have found that frame-wise autoregressive decoding is the most effective, rather than row-based autoregressive decoding. The frame-wise AR decoding also aligns with the streaming properties of video modality.
    \item Second, we discovered the impact of the mask ratio is minimal, which contrasts obviously with the conclusions of most previous studies. This could be due to the autoregressive nature, which inherently blocks the flow and exchange of reverse information. Benefiting from the ability to operate without an excessively large mask ratio, our method exhibits superior sample efficiency.
    \item Third, to leverage stronger generalization capabilities, we follow UMT~\cite{li2023unmasked} by aligning the autoregressive decoding features with CLIP~\cite{radford2021learning}, enabling fair comparisons with other baseline methods.
\end{itemize}
Next, we provide a detailed introduction to the hybrid encoder in VideoMAP and its frame-wise autoregressive alignment strategy.

\subsection{Hybrid Mamba-Transformer Video Encoder}
\label{sec:Hybrid Mamba-Transformer Video Encoder}





\begin{figure}[ht]
    \centering
    \includegraphics[width=0.5\linewidth]{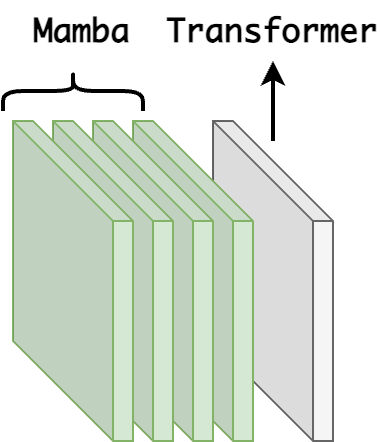} 
    \caption{We propose a 4:1 ratio Mamba-Transformer block.}
    \label{fig:hybridblock_image}
        \vspace{-5mm}
\end{figure}

In VideoMamba\cite{li2025videomamba}, the authors identified a significant overfitting issue in larger-scale models, which limited their scalability. While Transformers are well-known for their superior scalability, our finding is that inserting Transformer layers at regular intervals within Mamba layers can significantly improve model scalability, thereby achieving a balance between computational efficiency and performance. Consequently, a method that reduces GPU memory usage, enhances sample efficiency, and is easily scalable is critically important. We found that inserting a Transformer layer after every \(4N\) Mamba layers helps emphasize global details and mitigate the loss of global information, thereby making the Mamba-based backbone easier to scale up. To this end, we designed a hybrid Mamba-Transformer video encoder, referred to as \textbf{VideoHybrid}. This new architecture demonstrates that it is feasible to combine the strengths of Mamba and Transformer architectures in a hybrid manner, achieving significant improvements in scalability and sample efficiency. As illustrated in Figure~\ref{fig:hybridblock_image}, we employ a 4:1 Mamba-to-Transformer ratio to minimize additional computational overhead.

\subsection{Autoregressive Decoder for VideoMAP}
\label{sec:Autoregressive Decoder}
VideoMAP adopts a frame-wise autoregressive approach to align with CLIP features, thereby enhancing its temporal modeling capabilities. Compared to VideoMamba's UMT, which decodes all frame CLIP features simultaneously, our simple yet effective approach significantly improves the model's understanding of temporal dynamics, resulting in impressive advancements in capturing motion information and temporal correlations. 
The masking strategy followed by previous work strengthens the network's ability to model local features while enabling the inserted Transformer layers to enhance bidirectional modeling capabilities. 

\noindent\textbf{Autoregressive Decoding Strategies.}  
The goal of autoregressive (AR) pre-training is to learn high-quality conditional probability distributions that enable the model to generate new sequences based on previously generated content. 

\begin{figure}[ht]
    \centering
    \includegraphics[width=\linewidth]{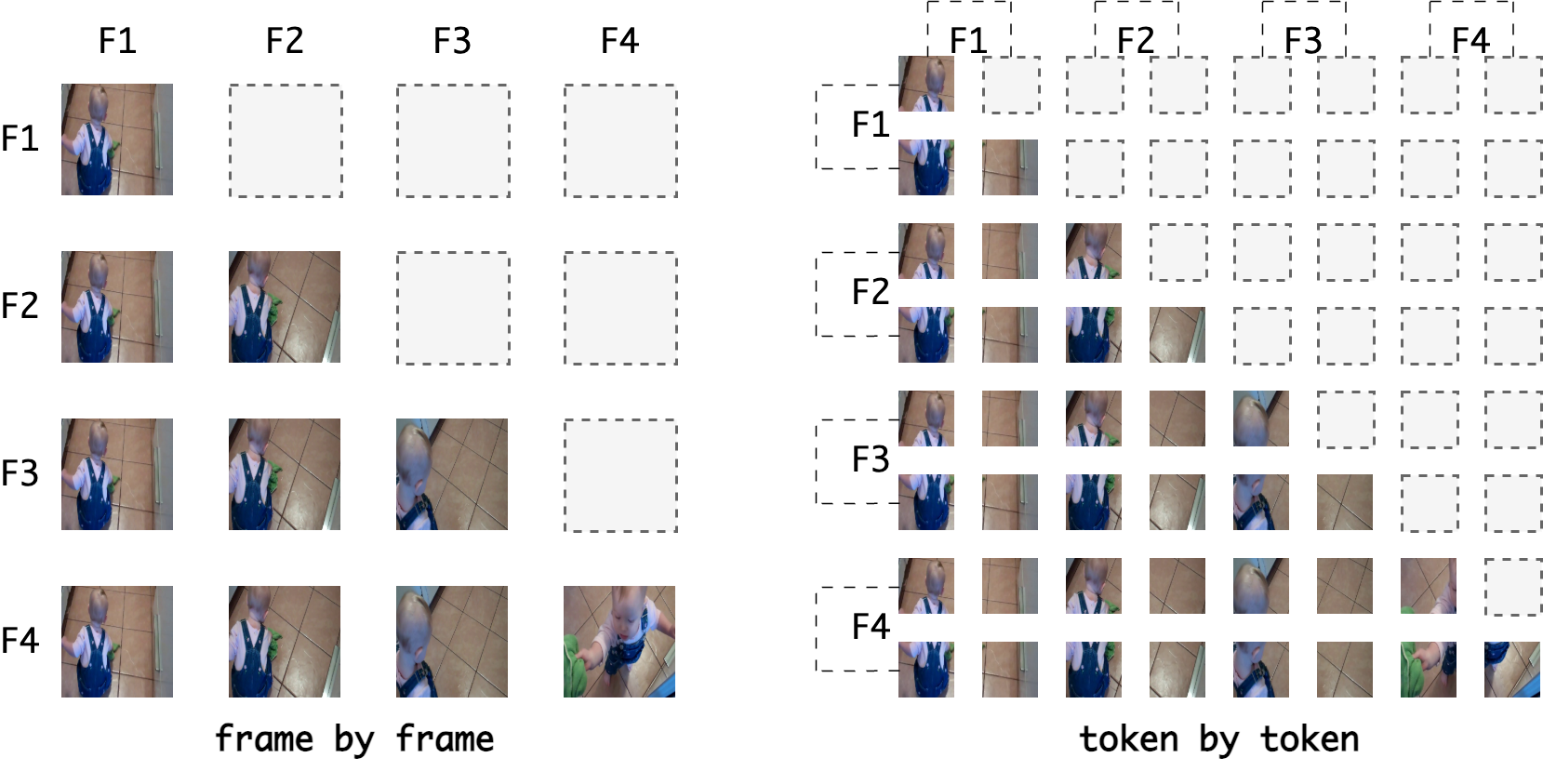} 
    \caption{We propose two autoregressive decoding strategies: \textbf{\textit{frame by frame}} and \textbf{\textit{token by token}}.}
    \label{fig:decoder_method}
        \vspace{-5mm}
\end{figure}

To achieve this, we first explore how the order of predictions in the autoregressive model influences the pre-training results of our VideoMAP framework. 
We propose two autoregressive pre-training decoding strategies for video, as illustrated in Figure~\ref{fig:decoder_method}.

The frame-by-frame method adopts an inter-frame autoregressive approach, where all tokens within a frame are visible to one another. The token-by-token method is an intra-frame autoregression strategy, where each token within a frame only has access to itself and the tokens that precede it. The results in Table~\ref{tab:decoder} show that both frame-wise and token-wise autoregressive decoding have demonstrated strong capabilities, but frame-wise autoregression achieves better top-1 accuracy. We attribute this to its enhanced global temporal modeling, so we adopt frame-by-frame autoregression as our default design.

\noindent\textbf{Decoder Block.}  
Our autoregressive decoder consists of three layers of Transformer blocks followed by one layer of MLP. And we remove the \texttt{cls} token to maintain consistency with the mask matrix.

To ensure the integrity of the information, we add the \texttt{cls} token back into the feature representation after the Transformer layers and use an MLP layer for mapping.

After processing through the decoder block, in order to train the model to predict the CLIP feature of the next frame, we compare the final output \( x_{\text{out}} \) with the target (i.e., the CLIP feature of the next frame). Specifically, we use Mean Squared Error (MSE) as the loss function to measure the difference between the two. 

By minimizing this loss function, the model gradually adjusts its parameters to better predict the CLIP features of the subsequent frame given the input. This training approach helps the model learn the temporal dynamics within video sequences, thereby improving the accuracy of its predictions for future frames. 

It is worth noting that during the encoder stage, the network performs bidirectional modeling, allowing information exchange between different frames for global feature extraction. Autoregression occurs only during the decoding alignment stage, where the network requires all historical global features and semantic information from the current frame to predict the semantic features of the next frame. This essentially demands that the network simultaneously perform future spatial and semantic predictions, significantly enhancing its spatiotemporal feature extraction capability. Compared to the UMT employed in VideoMamba, VideoMAP places greater emphasis on spatial-temporal semantics rather than spatial semantics, resulting in superior performance on video tasks.

\subsection{VideoLLM Based on VideoMAP}  
\begin{figure}[ht]
    \centering
    \includegraphics[width=\linewidth]{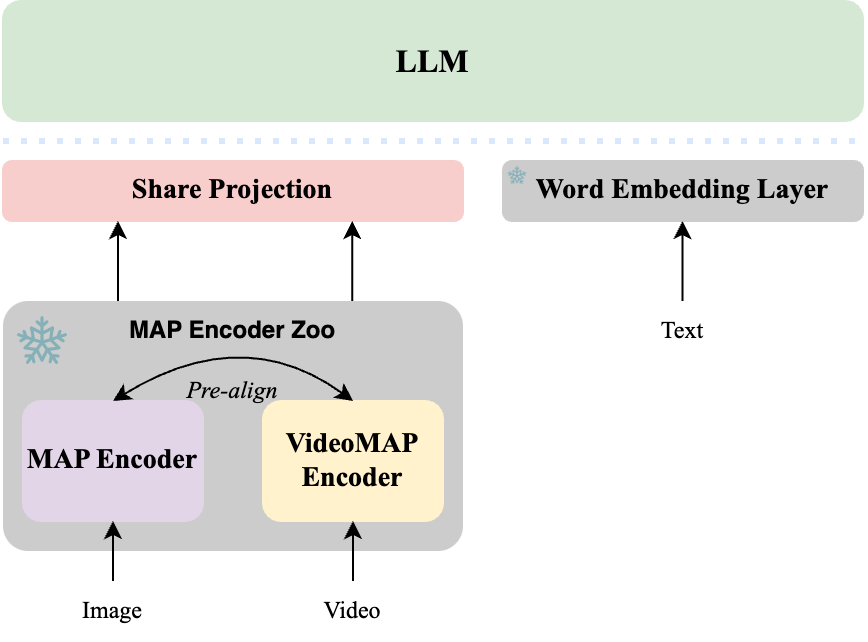} 
    \caption{We propose a VideoLLM based on VideoMAP and MAP and significantly reduce the amount of GPU memory.}
    \label{fig:videoLLM}
    \vspace{-5mm}
\end{figure}

To showcase the potential of VideoMAP in the VideoLLM domain, we conducted a fair comparison with methods based solely on Transformers and solely on Mamba, such as UMT and VideoMamba.
As illustrated in Figure~\ref{fig:videoLLM}, we used the framework of LLaVA~\cite{liu2024visual} and adopted the pre-training strategy and parameters similar to those in Video-LLaVA~\cite{lin2023video}. The first stage involves our MAP and VideoMAP visual alignment, where both images and videos are included in a batch. These are subsequently converged into a unified visual feature through a shared feature space that represents both images and video.

By pre-aligning the inputs to the LLM, the gap between different visual signal representations is significantly reduced. Ultimately, we obtain a unified visual representation, which, after passing through the shared projection layer, is fed into the LLM. 

The second stage is understanding training, where the model learns to interpret visual signals across a wide variety of image and video-text pair datasets. Due to the use of the hybrid vision encoder framework, GPU memory consumption during training and inference is reduced. The final stage involves regular instruction tuning of our pre-trained models.

\section{Experiments}
\label{sec:experiments}
In this section, we conduct experiments on video understanding tasks to verify the effectiveness of VideoMAP. Then, we demonstrate the application potential of VideoMAP in videoLLMs. Finally, we perform various ablation studies to validate each design choice.

\subsection{Video Understanding Tasks}
\noindent\textbf{Datasets and Settings.} 
We pretrained and fine-tuned on the Kinetics-400~\cite{kay2017kinetics} and Something-Something V2~\cite{goyal2017something} datasets. We report top-1 and top-5 validation accuracies, where the image size is \(224 \times 224\). We further validated the effectiveness of VideoMAP on the Breakfast\cite{breakfast} and COIN\cite{coin} datasets for long-term video tasks.
In the pretraining phase, we adopted the training method from CLIP-B~\cite{radford2021learning} used in UMT~\cite{li2023unmasked} and VideoMamba~\cite{li2025videomamba} for feature alignment. Specifically, we employed an attention-masking strategy with a 50\% masking rate. We used AdamW as the optimizer, with the warmup epochs set to 40 and the initial learning rate \(lr\) of 1.5e-4, dynamically adjusted according to the total batch size. We performed supervised fine-tuning for 45 epochs and reported the results. We use the same hyperparameters as VideoMamba to ensure a fair comparison.

\begin{table*}[t]
\centering
\tiny
\renewcommand{\arraystretch}{1.2}

\resizebox{0.95\linewidth}{!}{
\begin{tabular}{ccccccc}
\hline
Arch. & Model & Input.size & \#Params(M) & Epoch & Top-1 & Top-5 \\
\hline
Trans & BEVT-B & \(32\times224^2\) & 88  & 800 & 81.1 & - \\
Trans & ST-MAE-B & \(16\times224^2\) & 87 & 1600 & 81.3 & 94.9 \\
Trans & VideoMAE-S & \(16\times224^2\) & 22 & 2400 & 79.0 & 93.8 \\
Trans &  VideoMAE-B & \(16\times224^2\) & 87 & 1600 & 81.5 & 95.1 \\
Trans & UMT-B & \(8\times224^2\) & 87 & 800 & {85.7} & 97.0 \\
\hline
SSM & VideoMamba-M* & \(8\times224^2\) & 74 & 100 & 80.7 & 94.5\\
SSM & VideoMamba-M & \(8\times224^2\) & 74 & 800 & 82.0  & 95.4 \\
SSM & VideoMamba-M & \(16\times224^2\) & 74 & 800 & 83.4 & 95.9 \\
SSM & VideoMamba-M & \(32\times224^2\) & 74 & 800 & 83.9 & 96.2 \\
SSM & VideoMamba-B* & \(8\times224^2\) &98  & 800 & 81.4 & 95.0 \\
\hline

 SSM+Trans & VideoHybrid-M+UMT* & \(8\times224^2\) & 96 & 800 & 84.5  & 96.5 \\
 SSM & VideoMamba-M+VideoMAP* & \(8\times224^2\) & 74 & 800 &  83.5 &  96.0 \\
 Trans & UMT-B+VideoMAP* & \(8\times224^2\) & 87 & 800 & 85.2  & 96.7 \\
\rowcolor{blue!10} SSM+Trans & VideoMAP-M & \(8\times224^2\) &  96 & 100 & 83.5 & 96.1 \\
\rowcolor{blue!10} SSM+Trans & {VideoMAP-M} & \(8\times224^2\) &  96 & 800 & {85.5} & {97.1} \\
\rowcolor{blue!10} SSM+Trans & {VideoMAP-M} & \(16\times224^2\) &  96 & 800 & {85.8} & {97.3} \\
\rowcolor{blue!10} SSM+Trans & {VideoMAP-M} & \(32\times224^2\) & 96 & 800 & {86.1}  & {97.4} \\

\hline
\rowcolor{blue!10} SSM+Trans & VideoMAP-B & \(8\times224^2\) &  171 & {100} & {85.5} & {97.0} \\
\rowcolor{blue!10} SSM+Trans & {VideoMAP-B} & \(8\times224^2\) &  171 & {800} & {86.9} & {97.5} \\
\hline
\rowcolor{blue!10}SSM+Trans& {VideoMAP-L} & \(8\times224^2\) & 301  & {100} & {87.3} & {97.6} \\
\rowcolor{blue!10}SSM+Trans& {VideoMAP-L} & \(8\times224^2\) & 301  & {400} & \textbf{88.3} & \textbf{98.3} \\
\hline
\end{tabular}}
\vspace{-3mm}
\caption{Comparison with the state-of-the-art methods on Kinetics-400. The asterisk denotes the results of other reproduced baselines. VideoMamba-M* refers to the reproduced results of VideoMamba after 100 epochs of pretraining. VideoMamba-B* represents the results obtained by further scaling up the model following \cite{li2025videomamba}. VideoHybrid-M+UMT* represents the pretraining of a hybrid architecture using UMT. VideoMamba-M+VideoMAP* signifies the pretraining of VideoMamba using VideoMAP, whereas UMT-B+VideoMAP* denotes the application of VideoMAP for pretraining a Transformer-based architecture.}
\label{table1}
\vspace{-1mm}
\end{table*}

\noindent\textbf{Results.}
Tables~\ref{table1} shows the results in Kinetics-400. We observed the following advantages of VideoMAP: 
\begin{itemize}
    \item Better scalability compared to VideoMamba: When VideoMamba is scaled up to 98M parameters, it experiences a performance drop compared to its 74M version (as also noted in its paper). By incorporating a small number of Transformer layers, VideoMAP enhances scalability and strengthens long-context modeling capabilities. Even when expanded to 301M parameters, VideoMAP still achieves a notable performance improvement.
    \item More suitable for pre-training Mamba-based networks: Compared with VideoHybrid-M+UMT that adopts a hybrid architecture but uses UMT for pre-training—VideoMAP-M achieves a 1\% higher top-1 accuracy. Furthermore, using VideoMAP to pre-train VideoMamba improves performance by 1.5\% compared to UMT, indicating that VideoMAP is better adapted to Mamba-based architectures.
    \item Balanced performance and computational overhead: Although adding a small number of Transformer layers increases computation slightly, our hybrid architecture still supports longer input sequences. We observe further performance gains at 16 and 32 frames, even surpassing purely Transformer-based architectures.
\end{itemize}

Moreover, VideoMAP outperforms VideoMamba after just 100 epochs of pretraining, highlighting the significant advantages in sample efficiency and training time. Table~\ref{table2} and table~\ref{results_breakfast_coin} also shows the results on Something-Something V2 datasets, Breakfast~\cite{breakfast}, and COIN~\cite{coin} datasets. Experimental results show that our method not only excels in short-term tasks but also remains effective for long-term video. We found that VideoMAP delivers better performance and can be easily scaled to larger model sizes, further demonstrating the effectiveness of our approach. 



\begin{table*}[t]
\centering
\tiny
\renewcommand{\arraystretch}{1.2}

\resizebox{0.9\linewidth}{!}{
\begin{tabular}{ccccccc}
\hline
Arch. & Model & Input.size & \#Params(M) & Epoch & Top-1 & Top-5 \\
\hline
Trans & BEVT-B & \(32\times224^2\) & 88 & 800 & 70.6 & - \\
Trans & VideoMAE-S & \(16\times224^2\) & 22 & 1600 & 66.8 & 90.3 \\
Trans &  VideoMAE-B & \(16\times224^2\) & 87 & 2400 & 70.8 & 92.4 \\
Trans & UMT-B & \(8\times224^2\) & 87 & 800 & 70.8 & 92.6 \\
\hline
SSM & VideoMamba-M & \(8\times224^2\) & 74 & 800 & 70.2  & 92.6 \\
SSM & VideoMamba-M & \(16\times224^2\) & 74 & 800 & 71.0 & 92.7 \\
\rowcolor{blue!10} SSM+Trans & VideoMAP-M & \(8\times224^2\) & 96 & 800 & 70.6 & 93.0 \\
\rowcolor{blue!10} SSM+Trans & VideoMAP-M & \(16\times224^2\) & 96 & 800 & {71.2} & {93.6} \\
\rowcolor{blue!10} SSM+Trans & VideoMAP-B & \(8\times224^2\) & 171 & 800 & \textbf{72.0} &  \textbf{94.0} \\

\hline

\end{tabular}}
\vspace{-3mm}
\caption{Comparison with the state-of-the-art on Something-Something V2. VideoMAP demonstrates scalability to larger model sizes while achieving enhanced performance.}
\vspace{-5mm}
\label{table2}
\end{table*}


\begin{table*}[ht]
\centering
\resizebox{0.8\linewidth}{!}{
\begin{tabular}{lcc|cc|cc|cc}
\hline
\multirow{2}{*}{Methods} & LLM & Vision Encoder & \multicolumn{2}{c|}{MSRVTT-QA} & \multicolumn{2}{c|}{TGIF-QA} & \multicolumn{2}{c}{ActivityNet-QA} \\
& Size & Type & Accuracy & Score & Accuracy & Score & Accuracy & Score \\
\hline
VideoMamba-LLaVA & 7B & Vim-B + VideoMamba-M & 22.1 & 1.1 & 12.7 & 0.7 & 16.9 & 0.9 \\
VideoMAE-LLaVA & 7B & ViT-B+UMT-B & 30.6 & \textbf{1.7} & \textbf{21.0} & \textbf{1.3} & 25.6 & 1.3 \\
\rowcolor{blue!10}VideoMAP-LLaVA & 7B & MAP-B + VideoMAP-M & \textbf{31.0} & \textbf{1.7} & {20.7} & \textbf{1.3} & \textbf{26.7} & \textbf{1.4} \\
\hline
\end{tabular}}
\vspace{-3mm}
\caption{Comparison between different visual encoder for VideoLLM on video reasoning benchmarks. We employ Qwen2.5 to evaluate the performance of each model. And we choose the 72B-Instruct version which is the Top-1 LLM model in CompassRank~\cite{2023opencompass} Large Language Model Leaderboard. The results show that while our VideoMAP visual encoder reduces memory requirements, it also outperforms UMT-based methods of the same model size.}
\label{Videollm}
\vspace{-1mm}
\end{table*}

\begin{table}[t]
    \vspace{-0.3cm}
    \centering
    \setlength\tabcolsep{2pt}
    \resizebox{0.95\linewidth}{!}{
        \begin{tabular}{l|c|l|l|l|cc}
        \hline
            \multirow{2}*{\textbf{Method}} & \multirow{2}*{\textit{\textbf{e2e}}} & \multirow{2}*{\textbf{Backbone}} & \multirow{2}*{\textbf{Neck Type}} & \textbf{Pretraining}  & \textbf{BF} & \textbf{COIN} \\
            ~ & ~ & ~ & ~ & \textbf{Dataset} & \textbf{Top-1} & \textbf{Top-1} \\
            \hline
             Timeception~\cite{timeception} & \xmark & 3D-ResNet & Conv. & IN-1K+K400 & 71.3 & - \\
             VideoGraph~\cite{videograph} & \xmark & I3D & Conv.+Atten. & IN-1K+K400 & 69.5 & - \\
             GHRM~\cite{ghrm} & \xmark & I3D & Graph Conv.. & IN-1K+K400 & 75.5 & - \\

             Distant Supervision~\cite{distant} & \xmark & TimeSformer & Atten. w/ KB & IN-21K+HTM & {89.9} & {90.0} \\
             ViS4mer~\cite{vis4mer} & \xmark & Swin-B & SSM & IN-21K+K600 & {88.2} & {88.4} \\
             \hline
             Turbo$_{f32}$~\cite{turbo} & \cmark & VideoMAE-B &  & K400 & 86.8 & 82.3 \\
             Turbo$_{f32}$~\cite{turbo} & \cmark & VideoMAE-B &  & K400+HTM-AA & {91.3} & {87.5} \\
              
             VideoMamba$_{f32}$ & \cmark & VideoMamba-Ti &  & K400 & 94.3 & 86.2 \\
              
             VideoMamba$_{f64}$ & \cmark & VideoMamba-Ti &  & K400 & 94.3 & 87.0 \\
              
             VideoMamba$_{f32}$ & \cmark & VideoMamba-S &  & K400 & 95.3 & 88.4 \\
              
             VideoMamba$_{f64}$ & \cmark & VideoMamba-S &  & K400 & 97.4 & 88.7 \\
              
             VideoMamba$_{f32}$ & \cmark & VideoMamba-M &  & K400 & 94.8 & 88.3 \\
              
             VideoMamba$_{f64}$ & \cmark & VideoMamba-M &  & K400 & 95.8 & 89.5 \\
              
             VideoMamba$_{f32}$ & \cmark & VideoMamba-M{$\dag$} &  & K400 & \textbf{97.9} & 89.6 \\
              
             VideoMamba$_{f64}$ & \cmark & VideoMamba-M{$\dag$} &  & K400 & 96.9 & {90.4} \\
             \rowcolor{blue!10} VideoMAP$_{f24}$ & \cmark & VideoMAP-M &  & K400 & \textbf{97.9} & \textbf{90.9} \\
        \hline
        \end{tabular}
    }
    \caption{\textbf{Comparison with the state-of-the-art on Breakfast and COIN.} 
    ``\textit{e2e}'' means end-to-end methods without exhausting feature extraction. ``$\dag$'' marks the backbone with masked pretraining.}
    \label{results_breakfast_coin}
    \vspace{-0.6cm}
\end{table}

\subsection{VideoMAP-LLaVA}
\begin{figure}
    \centering
    \includegraphics[width=\linewidth]{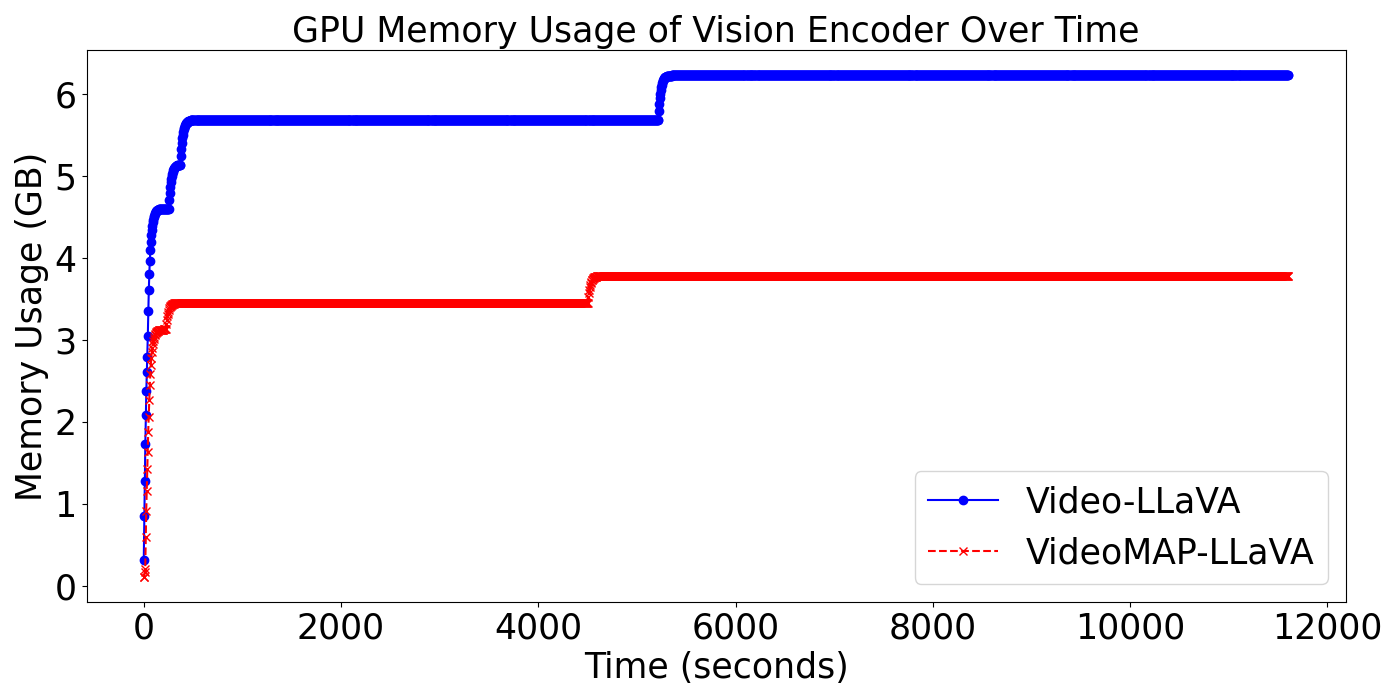}
    \vspace{-5mm}
    \caption{Comparison of the vision encoder on the same input. VideoMAP-LLaVA saves about 40\% of GPU memory.}
    \vspace{-7mm}
    \label{fig:gpu}
\end{figure}

In this section, to validate the ability of our VideoMAP as a visual encoder in the Multimodal Large Language Model domain, we constructed a Video-LLaVA style framework for two phases of training on the LLaVA~\cite{liu2024visual}, Valley~\cite{luo2023valley}, and Video-ChatGPT~\cite{Maaz2023VideoChatGPT} datasets, with pre-training and fine-tuning both performed for 1 epoch. 
Our approach employs MAP-B and VideoMAP-M as the image and video feature extractors, respectively. We also evaluated the results using Vim-B with VideoMamba-M and ViT-B with UMT-B as the visual extractors. This is a fair comparison, as all models are of similar size, and the datasets are limited to ImageNet and K400. 
Through the detailed recordings of the training and inference processes of VideoMAE-LLaVA and VideoMAP-LLaVA, we obtained detailed GPU memory usage comparison figures as shown in Figure~\ref{fig:gpu}.

\noindent\textbf{Comparisons of the GPU Memory for Vision Encoder.}
We can find in Figure~\ref{fig:gpu} that VideoMAP-LLaVA saves about 40\% of GPU memory for the vision encoder compared with the Transformer-based visual encoders. Since we adopt the same LLM backbone as Video-LLaVA, we only demonstrate here that our method saves memory costs in the vision encoder aspect. The LLM backbone part, being an FP16 type 7B model, occupies approximately 14G of memory. Replacing the LLM backbone with a hybrid framework to further save memory is also an important issue, although this is not within the scope of this research. We consider it an future work.

\noindent\textbf{Comparisons of the Performance.}
As shown in Table~\ref{Videollm}, we evaluate the VideoMAP-LLaVA on 3 video multimodal QA datasets. 
Compared with using VideoMamba-LLaVA and VideoMAE-LLaVA as visual encoders, VideoMAP-LLaVA achieves impressive results on MSRVTT\cite{msrvtt}, TGIF-QA\cite{jang2017tgif}, and ActivityNet-QA\cite{activitynet}. Here, we conduct a fair comparison of various visual encoder architectures in terms of both performance and computational cost. We believe that incorporating additional datasets will further enhance model performance. For instance, Video-LLaVA leverages a larger training corpus and extra modalities; its LanguageBind vision encoder is fine-tuned on VIDAL-10M (which includes 12 large-scale video datasets) to adjust CLIP, whereas our data source only utilizes ImageNet-1k and K400(230k) datasets, resulting in a significant difference in the volume of pre-training data. The Vision encoder of Video-LLaVA uses nearly 45 times the amount of data compared to what we use.  We believe that training our VideoMAP on larger datasets and integrating it with a hybrid Mamba-Transformer LLM backbone to form a new large language model is a valuable direction for future work.



\subsection{Ablation Study}
In this section, we conduct ablation studies to validate the rationality of the mask ratio used, the data efficiency of VideoMAP, and the frame-wise decoder design.

\noindent\textbf{Mask Ratio of Masked Pretraining.}
We used the attention masking mechanism to retain adjacent meaningful content and conducted experiments on three different ratios on the Kinetics-400 dataset. We employ VideoMAP-B, pretrain it on K400 for 100 epochs, and then report the results after fine-tuning. As shown in Table~\ref{table3}, we find that varying the masking ratio has minimal impact on fine-tuning results, suggesting that masking is not a critical factor in this setting. This observation notably differs from earlier approaches—such as VideoMAE and UMT—that rely heavily on larger mask ratio, and also contrasts with the smaller mask ratios used for image-based MAP. We retain masking in our approach and empirically set the default ratio to 50\%.

\begin{table}[ht]
\centering
\renewcommand{\arraystretch}{1.2}
\resizebox{\linewidth}{!}{
\begin{tabular}{ccccccc}
\hline
Model & Input.size & Mask Ratio & Top-1 & Top-5 \\
\hline
VideoMAP-B & \(8\times224^2\) & 0.2 & {85.37} & {96.89} \\
VideoMAP-B & \(8\times224^2\) & 0.5 & \textbf{85.54} & \textbf{96.96} \\
VideoMAP-B & \(8\times224^2\) & 0.8 & 85.21 & 96.89 \\

\hline
\end{tabular}}
\vspace{-3mm}
\caption{Ablation study on different mask ratios.}
\vspace{-3mm}
\label{table3}
\end{table}

\noindent\textbf{Performance Gain on Small Dataset.}
We constructed an ablation experiment on a small dataset by sampling 10\% samples in the Kinetics-400 training set. And we call this dataset K400-small. In K400-small, all experiments were pretrain 150 epochs and fine-tuned 25 epochs. As shown in Table~\ref{table4}, we performed detailed comparison experiments on VideoMamba, VideoHybrid-M+UMT, and VideoMAP in K400-small. VideoHybrid-M+UMT denotes the use of UMT to pre-train the hybrid framework VideoHybrid. 
Through the experiments in the K400-small dataset, we can see that VideoMamba shows a lack of generalization ability after a significant reduction in the data volume. 
VideoMAP achieves more significant performance improvements with smaller data volumes, demonstrating the data efficiency of VideoMAP. Compared to VideoHybrid-M+UMT and VideoMAP, both use the VideoHybrid network but adopt different pre-training strategies. It can be seen that the hybrid network pre-trained with VideoMAP significantly outperforms the one pre-trained with UMT by 2.6\%, which again confirms that VideoMAP performs better on the hybrid framework than the traditional UMT strategy used in Transformers.

\begin{table}[ht]
\centering
\renewcommand{\arraystretch}{1.2}

\resizebox{\linewidth}{!}{
\begin{tabular}{ccccccc}
\hline
Arch. & Model & Input.size & \#Params(M) & Dim & Top-1 & Top-5 \\
\hline
SSM & VideoMamba-M & \(8\times224^2\) & 74 & 576 & 55.5  & 89.7 \\
SSM+Trans & VideoHybrid-M+UMT & \(8\times224^2\) & 96 & 576 & 69.1 & {95.0} \\
\rowcolor{blue!10}SSM+Trans & VideoMAP-M & \(8\times224^2\) & 96 & 576 & {71.7} & 94.8 \\
\rowcolor{blue!10}SSM+Trans & VideoMAP-B & \(8\times224^2\) & 171 & 768 & \textbf{77.9} & \textbf{96.8} \\
\hline
\end{tabular}}
\vspace{-3mm}
\caption{Performance gain on small datasets. Our hybrid framework uses a mask ratio of 0.5, while VideoMamba follows the optimal parameter mask ratio of 0.8 they provide.}
\vspace{-3mm}
\label{table4}
\end{table}

\noindent\textbf{Autoregressive Decoder Strategy.}
To validate the rationality of our frame-wise alignment strategy, we conducted experiments comparing the results of frame-wise and token-wise alignment. We also compared these with the commonly used UMT strategy in Transformers, which can be considered a Video-wise alignment strategy. 
The results are presented in Table~\ref{tab:decoder}, showing that the frame-wise decoding strategy outperforms the token-wise decoding strategy, and both are significantly better than the UMT, a video-wise decoding strategy. This experiment further confirms the suitability of VideoMAP pre-training for the hybrid framework and the rationality of the frame-wise decoding approach. 

\begin{table}[ht]
\centering
\renewcommand{\arraystretch}{1.2}

\resizebox{\linewidth}{!}{
\begin{tabular}{cccccc}
\hline
Arch. & Model & Input.size  & Dim & Top-1 & Top-5 \\
\hline
SSM+Trans & VideoHybrid-M+UMT & \(8\times224^2\) & 576 & 69.1 & {95.0} \\
\rowcolor{blue!10}SSM+Trans & VideoMAP-M\(_{\textbf{\textit{token}}}\) & \(8\times224^2\) & 576 & {71.4} & \textbf{96.0} \\
\rowcolor{blue!10}SSM+Trans & VideoMAP-M\(_{\textbf{\textit{frame}}}\) & \(8\times224^2\) & 576 & \textbf{71.7} & 94.8 \\
\hline

\end{tabular}}
\vspace{-3mm}
\caption{Ablation study on different autoregressive strategies.}
\vspace{-5mm}
\label{tab:decoder}
\end{table}



\section{Conclusion}
\label{sec:conclusion}
This paper addresses the critical challenges in efficient video understanding, with a focus on scalability and data efficiency. We introduce VideoMAP, a Hybrid Mamba-Transformer framework that features a specialized pre-training approach designed to significantly mitigate overfitting. Experimental results demonstrate that VideoMAP outperforms existing models in both performance and scalability across various datasets and showcase the potential of VideoMAP as a visual encoder for multimodal large language models, emphasizing its ability to reduce memory usage and enable the processing of longer video sequences.

{
    \small
    \bibliographystyle{ieeenat_fullname}
    \bibliography{main}
}

\end{document}